# Self-Supervised Learning of Depth and Ego-Motion from Video by Alternative Training and Geometric Constraints from 3D to 2D

Jiaojiao Fang, Guizhong Liu

*Abstract*—Self-supervised learning of depth and ego-motion from unlabeled monocular video has acquired promising results and drawn extensive attention. Most existing methods jointly train the depth and pose networks by photometric consistency of adjacent frames based on the principle of structure-from-motion (SFM). However, the coupling relationship of the depth and pose networks seriously influences the learning performance, and the re-projection relations is sensitive to scale ambiguity, especially for pose learning. In this paper, we aim to improve the depth-pose learning performance without the auxiliary tasks and address the above issues by alternative training each task and incorporating the epipolar geometric constraints into the Iterative Closest Point (ICP) based point clouds match process. Distinct from jointly training the depth and pose networks, our key idea is to better utilize the mutual dependency of these two tasks by alternatively training each network with respective losses while fixing the other. We also design a log-scale 3D structural consistency loss to put more emphasis on the smaller depth values during training. To makes the optimization easier, we further incorporate the epipolar geometry into the ICP based learning process for pose learning. Extensive experiments on various benchmarks datasets indicate the superiority of our algorithm over the state-of-the-art self-supervised methods.

## I. INTRODUCTION

DYNAMIC 3D scene structure understanding is a key yet challenging problem in robotics and autonomous driving scenarios. Obtaining the accurate scene structure and objects' locations in the real world are essential for motion planning and decision making. The supervised methods require densely annotated ground-truth information from additional expensive sensors and precise calibration which are costly and time-consuming. Thus, the recent works seek to obtain the 3D scene geometric information and the ego-motion of the agents in a self-supervised manner from either stereo image pairs [10] [33] or video sequences [34].

The self-supervised learning framework that jointly optimizes the relative pose and the scene depth has caught the attention of academics as it depends much less on the ground-truth labels. Previous methods mainly rely on minimizing the image brightness consistency error among adjacent views by re-projecting the back-projected 3D points in the source views, which may contain much system error in realistic scenes due to the moving objects, repetitive textures, reflective surfaces, scale variations, and occlusions. Thus, some later works try to explicitly measure the inferred geometry of the whole scene by the 3D geometric alignment loss [22, 40] or the epipolar geometric loss [3, 41], which are important for self-supervised depth and pose learning. Although these frameworks have achieved excellent improvement, the following problems still exist: i) as the performance of the depth and pose estimations are inter-dependence, due to the scale ambiguity, the relations of the back-projection and re-projection would produce degenerated results, especially for the pose estimation. Thus more feasible optimization method is needed to obtain more reliable results. ii) The smaller depth values contain richer information and more important for depth estimation, while the large depth values are less important and always tolerated with larger estimation errors in a real application, thus the smaller depth value should be given a larger weight to avoid overly emphasizing the larger depth errors during training. iii) Although the ICP based geometric constraints can simultaneously optimize the depth and pose learning and less affected by the scale ambiguity, the optimization objective towards best-fit transformation seems tough to achieve and is in short of exact linear mathematical constraint relations.

Inspired by the existing excellent work, in this paper, we tackle the above issues by fusing the epipolar geometry into ICP registration to simplify the optimization process and obtain more reliable results, incorporating the log-scale 3D structural consistency loss for aligning depth with pose, in the self-supervised framework. To ensure each network is directly optimized towards the gradient descent direction and increase the convergence speed, we alternatively train the depth and pose networks to align depth with pose and align pose with depth by turns according to the epipolar geometric constraints fused ICP registration. Moreover, we also verify the effectiveness of the properties of the epipolar geometry, namely the low-rankness and the self-expression in union-of-epipolar-

The authors are with the School of Electronics and Information Engineering, Xi'an Jiaotong University, Xi'an 710049, China (e-mail: 995541569@qq.com; liugz@xjtu.edu.cn).

subspaces, for depth and pose learning.

Our main contributions are the following:

**Epipolar geometric constraints embedded the 3D structural consistency.** Multi-views geometric consistency is vitally important for the self-supervised depth and pose learning based on the structure-from-motion. The ICP-based geometric constraints can simultaneously optimize the depth and pose networks, and are less affected by the scale-ambiguity problem. While directly using the best-fitted transformation to optimize is difficult to get ideal results, due to the indirect relations of the pose network learning and ICP registration. In this paper, instead of constraining the best-fit transformation computed by ICP registration, we propose a more direct manner to optimize the pose network by the epipolar geometric constraint in consideration of the ICP registration, which is easier to optimize, and embedded the ICP registration into the depth and pose learning process, called the geometric constraints from 3D to 2D. By the combination of ICP registration and epipolar geometry, we can find a better registration for geometric consistency and effective optimization objective. Furthermore, we also verify the effectiveness of the properties of epipolar geometry, namely the low-rankness and self-expression in union-of-subspaces, which could serve as a global regularization and deal with the moving objects, for the depth and pose learning.

**A log-scale 3D structural consistency measurement**. It would be better to avoid excessive optimization of the relative larger errors caused by the larger depth values during the training process, which is with greater tolerance in practical applications. To this end, we use the log-scale mean squared error to measure the 3D structural consistency, which is aimed to solve the scale inconsistent problem over samples and improve the learning performance.

**Alternatively training the depth and pose networks with different losses.** Inspired by the two-step optimization process of the ICP method, to ease the training procedure and better utilize the mutual dependency of the two tasks, we proposed to alternatively train the depth and pose networks with different geometric constraints for aligning depth with pose and aligning pose with depth by turns. We verify the effectiveness of the alternative training in the self-supervised depth and pose learning framework.

We show that the proposed geometric constraints can be explicitly incorporated into the training process without breaking the simplicity of inference. The proposed framework is extensively evaluated on various datasets and have achieved the state-of-the-art performance.

## II. RELATED WORK

Monocular depth estimation based on deep neural networks from stereo image pairs or video sequences have achieved great advancement. The respective methods mainly fell into two sorts, the supervised methods and the self-supervised methods.

### A. Supervised Monocular Depth Estimation

Supervised monocular depth estimation refers to the problem setting that trains with vast ground-truth data. Due to the superiority of deep learning and the availability of ground-truth data, the supervised learning frameworks [1, 6, 21, 20, 17] acquired advanced accuracy. Eigen et al. [6] firstly proposed the depth estimation from a single image by training a network on sparse labels provided by LiDAR scans. Liu et al. [21] used a convolutional neural network (CNN) combined with the conditional random field to learn monocular depth. Karsch et al. [17] proposed a technique that automatically generates plausible depth maps from videos using non-parametric depth sampling. Several works tried to further improve the accuracy of supervised depth estimation by using more robust losses [1, 7, 20]. But the superior performance of these supervised methods usually relied on high quality and pixel aligned ground-truth depth data for training, which is challenging to gain in various real-world environments. Fu et al. [7] introduced a spacing-increasing discretization (SID) strategy for supervised depth estimation.

### B. Self-supervised Learning of Depth and Pose

Recently, many self-supervised methods have been proposed due to their less dependence on the ground truth depth data [42-46]. Self-supervised depth estimation methods mainly utilize the multi-view information from multiple cameras or video sequences by the methodology of structure-from-motion. Here we focus on the most related self-supervised monocular depth and pose estimation from videos.

Zhou et al. [34] first presented a joint learning of depth and ego-motion from unlabeled videos in the self-supervised manner with the static scene assumption. Yin et al. [32] added a refinement network to the depth and pose networks in for estimating the residual optical flow and used the forward-backward consistency to account for the moving regions. Mahjourian et al. [22] proposed an Iterative Closest Point (ICP) based differentiable 3D loss, which directly penalizes the inconsistencies of the estimated depths without relying on re-projection. Chen et al. [2] proposed a method integrating both the geometric information and the semantic information for scene understanding. Ranjan et al. [28] jointly learned the motion segmentation, optical flow, camera motion, and depth to obtain the complete geometric structure and motion of the scene. Godard et al. [11] proposed an effective minimum photometric loss and an analytical binary mask to deal with the occlusions excluding the invalid regions. Bian et al. [40] proposed a geometry consistency loss for scale-consistent predictions and a self-discovered mask for handling moving objects and occlusions. Shen et al. [16, 41] incorporated the epipolar geometry for more robust geometric constraints, while the pre-computed feature points matching is needed. Poggi et al. [26] focused on the uncertainty estimation for self-supervised monocular depth estimation and showed how this practice improves depth accuracy. Guizilini et al. [12] proposed a novel self-supervised monocular depth estimation method combining geometry with a novel network structure called PackNet. Guizilini et al. [13] adopted a novel network architecture using a pre-trained semantic segmentation network to guide the geometric representation learning in a self-supervised monocular learning framework.

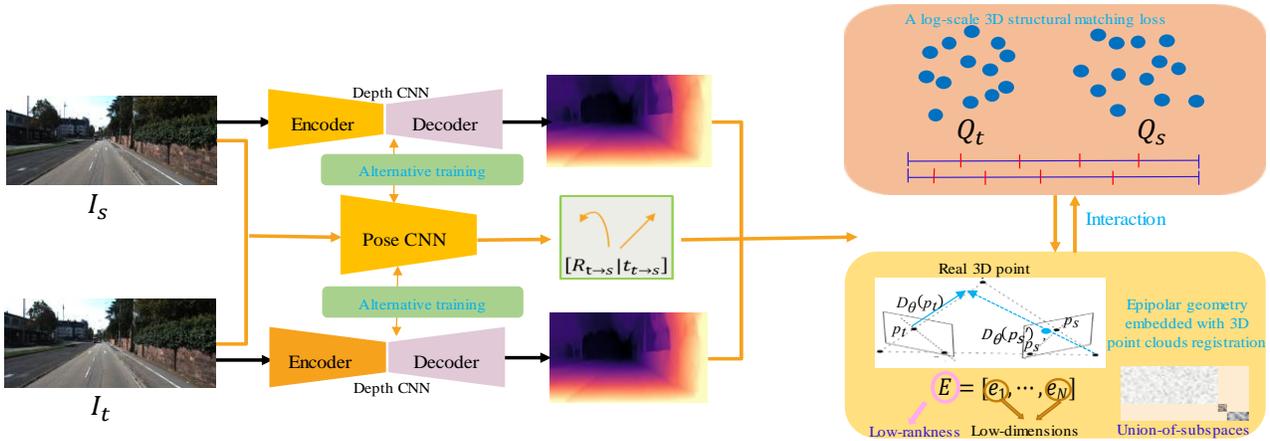

Fig. 1. An overview of our method. Besides photometric consistency, we explore the epipolar geometric consistency, the log-scale 3D structural consistency, and an alternative training policy with different geometric consistency for each network to improve the depth and pose estimation performance. Here the two depth CNNs have shared parameters.

*C. Epipolar Geometry for Self-supervised Learning*

The epipolar geometric constraints are popular for self-supervised optical flow, and depth-pose learning. Valgaerts et al. [35] introduced a model to simultaneously estimate the fundamental matrix and the optical flow. Wedel et al. [36] used a fundamental matrix as a weak constraint for the optical flow training. These methods, however, assumed that the scene was mostly rigid, and treated the dynamic parts as outliers [36]. Garg et al. [24] used the subspace constraint as a regularization term for multi-frame optical flow estimation. Zhong et al. [18] proposed a low-rank constraint as well as a union-of-subspaces constraint for the self-supervised optical flow training and investigated multiple ways of enforcing the epipolar constraint. While we explore to Chen et al. [3] captured multiple geometric constraints for relating the optical flow, depth, camera pose and intrinsic parameters from monocular videos, and used epipolar geometry as a verification of putative correspondences by optical flow.

In this work, we follow the previous good practices, with the major distinction that explore a new training policy to facilitate the training procedure of the depth and pose networks with a log-scale 3D structural consistency loss and the epipolar geometry and ICP based geometric constraint.

### III. THE PROPOSED METHOD

In this section, our proposed method is described, including the log-scale 3D structural consistency loss, the geometric constraints from 3D to 2D, the properties of the epipolar geometry as regularizations, as well as the alternative training with different losses. Fig. 1 illustrates an overview of our method. This section starts with an introduction to the problem formulation.

*A. Problem Formulation*

The problem of the self-supervised depth and pose networks learning from monocular video sequences can be formalized as follows. Given a target frame $I_t \in R^{H \times W \times 3}$ and the source frames $I_s$ where $s \in \{t-1, t+1\}$, collected by a potentially moving camera with intrinsic $K$. The 3D rigid transformation can be represented as a matrix $T_{s \to t} = [R_{s \to t}, t_{s \to t}]$, where $R_{s \to t}$ is the rotation matrix and $t_{s \to t}$ is the translation vector of the camera's ego-motion from time $s$ to $t$, which is to be estimated by the pose network. Assuming a pixel $p_t = [u_t, v_t]$ in the target frame $I_t$ and the estimated depth $D_\theta(p_t)$ of $p_t$, then the corresponding 3D locations $Q_t(p_t)$ in the camera's world coordinates system at time $t$ is the back-projection of $p_t$:

$$Q_t(p_t) = \begin{bmatrix} X_t \\ Y_t \\ Z_t \\ 1 \end{bmatrix} = D_\theta(p_t) K^{-1} \begin{bmatrix} u_t \\ v_t \\ 1 \end{bmatrix} \quad (1)$$

where $D_\theta$ represents the output of the depth network and $\theta$ refers to the parameters of the depth network.

Assuming $Q_t(p_t)$ is transformed rigidly from time $t$ to $s$, then its correspondence $p_s'$ in the source frame $I_s$ can be calculated by re-projection the back-projected 3D points $Q_t(p_t)$.

$$[p_s', 1]^T = D(p_t)^{-1} K T_{s \to t} Q_t(p_t) \quad (2)$$

where $D(p_t)$ is the depth of the transformed $Q_t(p_t)$ calculated by $T_{s \to t}$ in the camera's world coordinates at time $s$. Note that the transformation applied to the 3D point is the inverse of the camera movement from $t$ to $s$. Thus, the photometric loss between the target and source views can be easily computed by this displacement.

In this paper, we use the minimum error among adjacent views [11] as the photometric loss

$$\mathcal{L}_{ph} = \min_s [ph(I_t, I_{s \to t})] \quad (3)$$

where $I_{s \to t}$ is the reconstructed target image from the source images by the re-projection relation in Eq. (2). Following [10], the bilinear sampling [38] is used to reconstruct the target image, and the convex combination of L1 and structured similarity (SSIM) [39] to construct the photometric error $ph(\cdot,\cdot)$.

$$ph(\cdot,\cdot) = \alpha(1 - SSIM(\cdot,\cdot))/2 + (1 - \alpha)\|\cdot - \cdot\|_1 \quad (4)$$

As in [11], the edge-aware smoothness is used for depth maps

$$\mathcal{L}_s = |\partial_x D_\theta(p_t)^*| e^{-|\partial_x I_t|} + |\partial_y D_\theta(p_t)^*| e^{-|\partial_y I_t|} \quad (5)$$

where $D_\theta(p_t)^* = D_\theta(p_t)/\overline{D_\theta(p_t)}$ is the mean-normalized inverse depth to prohibit shrinking of the estimated depth [11].

A per-pixel binary mask $\mu$ proposed by Godard et al. [11] is used to exclude the regions which are harmful to the photometric loss. Thus the method in [11] trained on the monocular videos with loss $\mu\mathcal{L}_{ph} + \mathcal{L}_s$ is the baseline of our method.

*B. 3D Structural Consistency in Adjacent Views*

The scene structures obtained by depth estimations of adjacent views should be consistent [3]. Thus, enforcing the 3D structural consistency is necessary, which considers the whole scene structure. In this paper, to increase the importance of the smaller depth values and to avoid excessive optimization of the larger depth values, we introduce a log-scale 3D geometric consistency loss to penalize the structural variations in multiple views and enforce the scale consistency.

Given the relation between a pixel $p_t$ in the target image, and its correspondence $p_s'$ in the source image obtained by re-projection in Eq. (2), this relation can be used to penalize the structural inconsistency of adjacent views in the uniform coordinates system by a bilinear sampling [38]. In consideration of the occlusions, inspired by [11], we use the minimum error instead of the average error over all source views as the 3D geometric loss.

$$\mathcal{L}_{3D} = \min_s(\|\log(D_\theta(p_s')K^{-1}p_s' - \log(T_{s\to t}Q_t(p_t))\|) \quad (6)$$

By this way, the gradients of the larger depth value would be smaller during training. This loss is similar to the geometric consistency loss in [40] that computing the relative error instead of the absolute error while focus more on the smaller depth errors. And the experiments prove the effective of the simple logarithm operation.

*C. Epipolar Geometry Embedded with 3D Structural Consistency*

The epipolar geometric constraint is less affected by the depth estimation as it does not concern the depth in its formulation. Existing methods use either the pre-computed sparse feature matching [16] or the optical flow [3] combined with the estimated camera pose to construct the epipolar geometric constraint on the image planes, and improved the effectiveness of epipolar geometry.

In this paper, instead of using the optical flow, feature matching, or the re-projection of the back-projected 3D structure by the estimated depth of one image, we use the precise point clouds matching methods, called ICP, to establish the dense correspondences of the two sets of points $Q_t = \{Q_t(p_1), \cdots, Q_t(p_N)\}$ and $Q_s = \{Q_s(p_1), \cdots, Q_s(p_N)\}$ from adjacent views, where N is the total number of points. ICP method alternatively computes correspondences between two sets of 3D point clouds by searching the minimum point-to-point distances of each point pairs, and computing a best-fit transformation between the two sets of the point clouds with these correspondences. The next iteration then re-computes the correspondence with the previous iteration's transformation applied. Thus the optimization objectives of the n-th iteration are as follows:

$$\min_{s''\epsilon\{1,\cdots,N\}} \|Q_s(p_{s''}) - T'_{s\to t}Q_t(p_t)\|, t\epsilon\{1,\cdots,N\}. \quad (7)$$

Where $p_t$ and $p_{s''}$ denotes the point to point correspondence found by the closest point distances of the two sets of point clouds, $T'_{s\to t}$ represents the estimated pose of the (n-1)-th iteration. With the correspondences of the point clouds, the transformation between two adjacent views can be optimized by

$$\operatorname*{argmin}_{T_{s\to t}} \sum_{t=1}^{N} \|Q_s(p_{s''}) - T_{s\to t}Q_t(p_t)\|_2^2. \quad (8)$$

As the 3D structural consistency loss in Eq. (6), which is similar to [3, 40], is already an effective manner to obtain the 3D structural consistency of the estimated depth. Here instead of using the matching of point clouds reconstructed by the estimated depths of adjacent views to further constrain the 3D structural consistency of the adjacent views as in [22], we just use the matching of the two sets of point clouds to indicate the pixel relations of two adjacent views. To better utilize the mutual dependency of the depth and pose networks and to take the 3D geometric structural consistency into account, we embedded the correspondences of the two sets of point clouds into the 2D image planes. Instead of directly computing a transformation by the corresponding point clouds, we use the outputs of the pose network and the correspondences embedded in the two adjacent views to construct an epipolar geometry, which could transfer the ICP optimization into a linear problem. The optimization aims to transfer the consistency of depth to pose and constrain to consistent with epipolar geometry.

The correspondence between adjacent views in the image planes based on the ICP alignment are:

$$p_{s''} = p_t + [\delta_x, \delta_y]. \quad (9)$$

Assuming a pinhole imaging model, the correspondence, $p_t$ and $p_s''$, in adjacent views should satisfies the relation of the epipolar geometry with a fundamental matrix $F = K^{-1^T}[t_{s\to t}]_\times R_{s\to t}K^{-1}$, which are the cross product of the rotation matrix and the translation vector multiplied with the camera intrinsic.

$$\operatorname*{argmin}_{T_{s\to t}} \sum_{t=1}^{N}[p_{s''},1]F[p_t,1]^T, s.t. \min_{s''\epsilon\{1,\cdots,N\}} \|Q_s(p_{s''}) - T'_{s\to t}Q_t(p_t)\|, t\epsilon\{1,\cdots,N\} \quad (10)$$

where $[\cdot]_\times$ is the skew-symmetric matrix of a 3-dimensional vector. The obtained correspondences in Eq. (9) are integer-valued not needing any bilinear sampling. The roles of this loss can be two folds, one is a validation for the multi-views 3D structural consistency with an exact linear mathematical relation, and the other is for the pose network optimization.

The basic epipolar geometry is an over-constrained formulation, which are not robust to outliers or noises. To improve the robustness of the epipolar geometric constraint, in this paper we incorporate the low-rankness loss $\mathcal{L}_{lr}$ and self-expression in union-of-subspaces proposed in [18], into the self-supervised depth and pose learning.

To introduce the properties of the epipolar geometry, we first rewrite the epipolar geometric constraint in Eq. (10) as

$$f^T vec([p_s',1]^T[p_t,1])=0 \quad (11)$$

where $f$ is the vectorized fundamental matrix $F$ along the column direction, and $e_t = vec([p_s',1]^T[p_t,1])$ is the vectorized multiplication of the two points, which is a 9-dimensional column vector lying in the epipolar subspace [4]. Then we can form a matrix $E = [e_1, \cdots, e_N]$ by all the vectors $e_t$, where $t = 1: N$ is the number of pixel points. Thus the low-rankness loss of the matrix $E$ can be formulated by the nuclear

norm
$$\mathcal{L}_{lr} = \|E\|_* \tag{12}$$
where $\|\cdot\|_*$ is the nuclear norm, which can be computed by the singular value decomposition (SVD) [18]. By using this constraint, it is not necessary to explicitly compute the fundamental matrix. So it can be applied in the degenerated cases where a fundamental matrix is unknown. Although the low-rankness constraint is too loose, it still can improve the performance to some extent [18].

To deal with the moving objects in the static scene, another constraint, called the self-expression in union-of-subspaces, is introduced. This constraint implies that all vectors lying in the union-of-subspaces can be characterized by the self-expression property [18], i.e., each vector can be represented as a linear combination of the other vectors. The coefficients would be nonzero among the vectors in the same subspace while keeping zero among the vectors in others. The mathematical expression is [18]:
$$\mathcal{L}_{sub} = \frac{1}{2}\|(I + \lambda E^T E)^{-1}\lambda E^T E\|_F^2 + \frac{\lambda}{2}\|E(I + \lambda E^T E)^{-1}\lambda E^T E - E\|_F^2 \tag{13}$$
where C is the coefficients matrix of the self-expression in the union-of-subspaces, $\lambda = 0.05$ is a relaxing factor.
In consideration of the GPU memory and the computational efficiency, we randomly sample 2000 point pairs to compute this loss as in [18]. Even though these epipolar subspaces would be disjoint, it can still serve as a global regularization.

*D. Alternative Training by Different Optimization Objectives*

The ICP methods alternates between computing correspondences between two 3D point clouds by a simple closest point heuristic, and computing a best-fit transformation between the two point clouds with a given correspondences until convergence. We embed this traditional optimization idea into the self-supervised depth and camera pose learning framework, which align the 3D structure of the adjacent views with the estimated pose and learn the transformation between adjacent views by the pose network with a given correspondence from two sets of 3D points.

In this section, we describe the proposed training policy and losses used in the self-supervised depth and pose networks learning. To ensure that each of the networks is directly optimized towards the gradient descent direction, here we use an alternative training policy with different geometric constraints, the log-scale 3D geometric loss and the structural consistency embedded epipolar geometric constraint, and the properties of the epipolar geometry as regularizations for their learning respectively.

**Pose Network Optimization.** As the epipolar geometric constraints is less influenced by the depth estimation, it is a well way for pose estimation. Thus, in this paper we use the epipolar geometric constraint in Eq. (10) for better pose learning to ensure that each network is optimized towards the gradient descent direction. Here we use the epipolar geometric constraints in Eq. (8) to further constrain the pose network learning. Hence, we still use the minimum error to construct the epipolar geometric loss.
$$\mathcal{L}_{ep} = \min_s([p_s'', 1]F[p_t, 1]^T) \tag{14}$$
The epipolar geometric loss $\mathcal{L}_{ep}$ can be easily obtained by computing the distance map from each pixel to its corresponding epipolar line, as in [41]. As the camera pose estimation is more vulnerable to the moving objects in the static scene, the properties of self-expression in the union-of-subspaces $\mathcal{L}_{sub}$ proposed in [18] is also used to regularize the pose network optimization by the relation of the re-projection in Eq. (2). Thereby, the consistency between the two kinds of correspondences $p_s'$ and $p_s''$ would be guaranteed. Then the total loss of the pose network training is as follows:
$$\mathcal{L}_{pose} = \mu\mathcal{L}_{ph} + \lambda_s\mathcal{L}_s + \lambda_e\mathcal{L}_{ep} + \lambda_{su}\mathcal{L}_{sub} \tag{15}$$
where $\lambda_s$, $\lambda_e$, and $\lambda_{su}$ are the weights for the different losses.

**Depth Network Optimization.** The 3D geometric structural consistency loss can directly measure the 3D structure of the whole scene. Thus a 3D geometric loss in Eq. (6) is a better manner for directly optimizing the depth network. In consideration of the effectiveness of the low-rankness constraints $\mathcal{L}_{lr}$ reported in [18], here we also use it to regularize the depth learning. Thus the total loss for the depth network can be expressed as
$$\mathcal{L}_{depth} = \mu\mathcal{L}_{ph} + \lambda_s\mathcal{L}_s + \lambda_{3d}\mathcal{L}_{3D} + \lambda_l\mathcal{L}_{lr}, \tag{16}$$
where $\lambda_s$, $\lambda_{3d}$, and $\lambda_l$ are the weighs for the different losses.

These losses average over all the pixels, scales, and batches to train the networks in an end-to-end manner. The role of the photometric loss here is as the raw data term and as the global verification. To ease the training process and improve the learning performance, we alternatively train one network while fixed the other by the respective losses, which means that only the feed-forward will be performed for the other network without back-propagation.

IV. EXPERIMENTS

In this section, we evaluate the performance of our models and compare it with the published state-of-the-art self-supervised methods on the KITTI 2015 stereo dataset [9]. We also use the Make3D dataset [27] to evaluate the generalization ability on cross dataset.

*A. Training Datasets*

**KITTI Raw Dataset.** We mainly used the raw KITTI dataset [9] for training and evaluation. The dataset contains 42,382 rectified stereo pairs from 61 scenes, with a typical resolution being 1242×375 pixels. We trained our model with the Eigen split [6] excluding 679 images for testing and removed static frames following Zhou et al. [34]. This led to a total of **44,234** sequences, out of which we used **39,810** for training and **4,424** for validation. To facilitate the training and provide a fair evaluation, input images were resized to 640×192.

**KITTI Visual Odometry (VO) Dataset.** The KITTI odometry dataset contains 11 driving sequences with ground-truth labels and 11 sequences without ground-truth. As in the standard setting, we used sequences 00-08 for training and sequences 09 and 10 for testing.

**Cityscapes Dataset.** Since starting from a pre-trained model boosted the performance [34], we also tried to pre-train the model on the Cityscapes [5] dataset where 88084 images for training and 9659 images for validation.

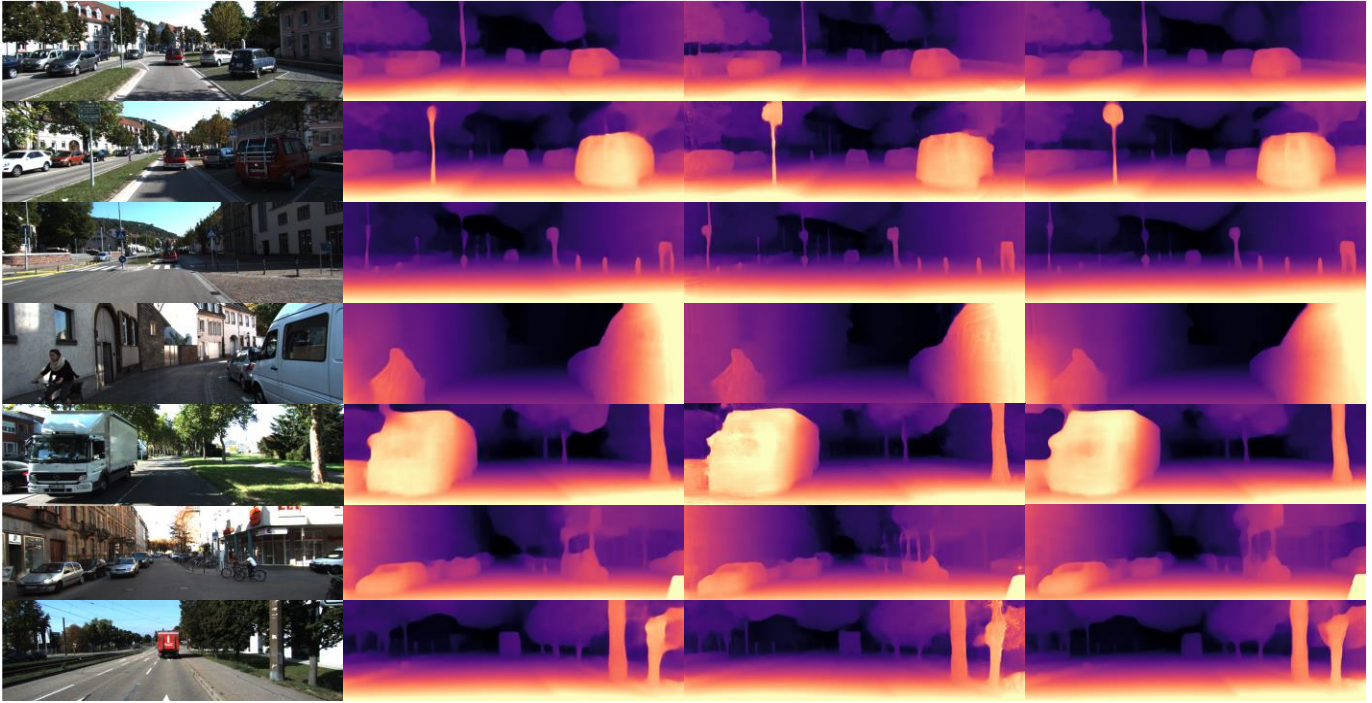

Fig. 2. The qualitative results of our proposed architecture on the KITTI dataset with the Eigen split [6]. The columns from left to right show respectively input images, the state-of-the-art predicted depth maps (Godard et al., 2019 [11]; Xue et al., 2020 [37]), and the depths maps obtained by our proposed architecture. Our method recovers more subtle details such as trees, trunks and advertising boards.

TABLE I
RESULTS OF COMPARISON WITH THE STATE-OF-THE-ART METHODS ON THE KITTI DATASET [9] WITH THE SPLIT OF EIGEN [6], WHERE THE ERROR METRIC IS LOWER THE BETTER, AND THE ACCURACY METRIC IS HIGHER THE BETTER. THE BEST RESULTS ARE IN BOLD; THE SECOND BEST IS UNDERLINED. 'K' REPRESENTS KITTI RAW DATASET AND 'CS' REPRESENTS CITYSCAPES TRAINING DATASET. M REFERS TO METHODS THAT TRAIN USING MONOCULAR SEQUENCES, S REFERS TO METHODS THAT TRAIN USING STEREO PAIRS, D REFERS TO METHODS THAT USE GROUND-TRUTH DEPTH SUPERVISION, 'SEM' REFERS TO METHODS THAT INCLUDE SEMANTIC INFORMATION. 'P-P' REFERS THE RESULTS OBTAINED WITHOUT POST-PROCESSING.

| Method | Supervision | Dataset | Error metric | | | | Accuracy metric | | |
|---|---|---|---|---|---|---|---|---|---|
| | | | Abs Rel | Sq Rel | RMSE | RMSE log | $\delta < 1.25$ | $\delta < 1.25^2$ | $\delta < 1.25^3$ |
| Depth capped at 80m | | | | | | | | | |
| Eigen et al. Fine[6] | D | K | 0.203 | 1.548 | 6.307 | 0.282 | 0.702 | 0.890 | 0.958 |
| Liu et al. [21] | D | K | 0.201 | 1.584 | 6.471 | 0.273 | 0.678 | 0.898 | 0.967 |
| Zhou et al.[34] | M | CS+K | 0.198 | 1.836 | 6.565 | 0.275 | 0.718 | 0.901 | 0.960 |
| Godard et al.[10] | S | CS+K | 0.141 | 1.186 | 5.677 | 0.238 | 0.809 | 0.928 | 0.969 |
| Vid2depth [22] | M | CS+K | 0.159 | 1.231 | 5.912 | 0.243 | 0.784 | 0.923 | 0.970 |
| Yin et al. [32] | M | CS+K | 0.149 | 1.060 | 5.567 | 0.226 | 0.796 | 0.935 | 0.975 |
| Shen et al. [16] | M | CS+K | 0.139 | 0.964 | 5.309 | 0.215 | 0.818 | 0.941 | 0.977 |
| Zhan et al.[33] | S | K | 0.135 | 1.132 | 5.585 | 0.229 | 0.820 | 0.933 | 0.971 |
| Monodepth2.[11] | M | K | 0.112 | 0.851 | 4.754 | 0.190 | 0.881 | 0.960 | 0.981 |
| Monodepth2. w/o p-p [11] | M | K | 0.115 | 0.903 | 4.863 | 0.193 | 0.877 | 0.959 | 0.981 |
| Chen et al. [3] | M | K | 0.135 | 1.070 | 5.230 | 0.210 | 0.841 | 0.948 | 0.980 |
| Guizilini et al.[13] | M+Sem | CS+K | 0.117 | 0.854 | 4.714 | 0.191 | 0.873 | **0.963** | 0.981 |
| Xue et al. [37] | M | K | 0.113 | 0.864 | 4.812 | 0.191 | 0.877 | 0.960 | 0.981 |
| Ours w/o p-p | M | K | 0.112 | 0.835 | 4.748 | 0.189 | 0.878 | 0.960 | 0.981 |
| Ours | M | CS+K | **0.110** | **0.793** | **4.674** | **0.186** | **0.884** | **0.963** | **0.982** |
| Ours | M | K | **0.110** | 0.806 | 4.681 | 0.187 | 0.881 | 0.961 | **0.982** |
| Depth capped at 50m | | | | | | | | | |
| Vid2depth.[22] | M | CS+K | 0.151 | 0.949 | 4.383 | 0.227 | 0.802 | 0.935 | 0.974 |
| Yin et al. [32] | M | CS+K | 0.147 | 0.936 | 4.384 | 0.218 | 0.810 | 0.941 | 0.977 |
| Shen et al. [16] | M | K | 0.133 | 0.778 | 4.069 | 0.207 | 0.834 | 0.947 | 0.978 |
| Ours | M | K | **0.105** | **0.616** | **3.602** | **0.179** | **0.890** | **0.966** | **0.984** |

*B. Implementation Details*

**Depth Network.** The depth network was a fully convolutional encoder-decoder structure with skip connections, which was similar to the DispNetS [23]. The ResNet18 [14] was used as the encoder if no otherwise specified. The decoder had five deconvolution layers. Networks outputted the results at 4 different spatial scales. The lower resolution depth maps were up-sampled to the input resolution for photometric loss as in [11].

**Pose Network.** The pose network took two adjacent frames as input, and outputted the relative motions between the target view and source views. The network consisted of 7 convolutional layers followed by a $1 \times 1$ convolution with 6 outputs channels, corresponding to rotation angles and translations along the coordinate axis.

TABLE II
ODOMETRY RESULTS ON THE KITTI [9] ODOMETRY DATASET. RESULTS SHOW THE AVERAGE ABSOLUTE TRAJECTORY ERROR AND STANDARD DEVIATION IN METERS.

| Methods | Sequence 09 | Sequence 10 | # frames |
|---|---|---|---|
| Garg et al. [8] | 0.013 ±0.010 | 0.012 ±0.011 | 3 |
| Zhou et al. [34] | 0.021 ±0.017 | 0.020 ±0.015 | 5 |
| Vid2depth[22] | 0.013 ±0.010 | 0.012 ±0.011 | 3 |
| GeoNet[32] | 0.012 ±0.007 | 0.012 ±0.009 | 5 |
| Ranjan et al. [28] | 0.012 ±0.007 | 0.012 ±0.008 | 5 |
| Monodepth2[11] | 0.017 ±0.008 | 0.015 ±0.010 | 2 |
| Shen et al. [16] | 0.009 ±0.005 | 0.008 ±0.007 | 3 |
| Ours | 0.008 ±0.005 | 0.008 ±0.006 | 3 |

**Parameters Setting and Processing.** For all the experiments, we set the weights of the different losses components as $\lambda_s = 0.001$, $\lambda_e = 0.002$, $\lambda_{3d} = 0.02$, $\lambda_l = 0.001$ and $\lambda_{su} = 0.0001$. We trained our model with the Adam [19] optimizer with $\beta_1 = 0.9$, $\beta_2 = 0.999$, Gaussian random initialization, ResNet18 with pre-trained weight on the ImageNet [29] and mini-batch size of 4. The learning rate was originally set to 0.0001 and halved it after every 10 epochs until the end. We used the same data augmentations as in [11]. For disparity maps, we followed a similar post-processing technique as in [11] and capped depth at 80m as per standard practice during evaluation [10].

**Equipment and Efficiency.** The algorithm was deployed in the PyTorch [25] framework which was compiled with CUDA 9.0 and CuDNN 7.0 on a computer with an Intel Xeon(R) E5-1660v4 HP-Z440 8-Core 3.2 GHz CPU and a Titan Xp GPU. With a single Titan Xp, the network took almost 3.1 hours per epoch compared with 0.8 hours of the baseline method. While the runtime of our model for testing was the same as the baseline.

If there was no additional specification, the models were trained by these conditions.

### C. Main Results of Depth and Pose Evaluations

**Depth evaluation.** The evaluation of depth estimation followed the previous works [11, 34, 22]. Here we provided a comparison of the depth estimation with the state-of-the-art self-supervised methods [3, 10, 11, 13, 16, 22, 33, 34, 37] and the classical supervised methods [6, 21]. To be fair for all methods, we used the same crop manner as [34] and evaluated the prediction with the same resolution as the input image. The measure criterion conformed to the one used in [11]. As shown in Table 1, with the same underlying network structure, the proposed method outperformed state-of-the-art methods by a large margin. The network first pre-trained on the larger Cityscapes dataset [5], and then fine-tuned on the KITTI dataset [9], would result in slight performance improvement. The final post-processing step led to an accuracy increase and fewer visual artifacts at the expense of doubling the test time. To prove the performance of the close-range depth estimation, we also provided the separate results for a depth capped at 50m, which was also shown the advantage of our method. Qualitative results compared with the predictions of Godard et al. [11] and

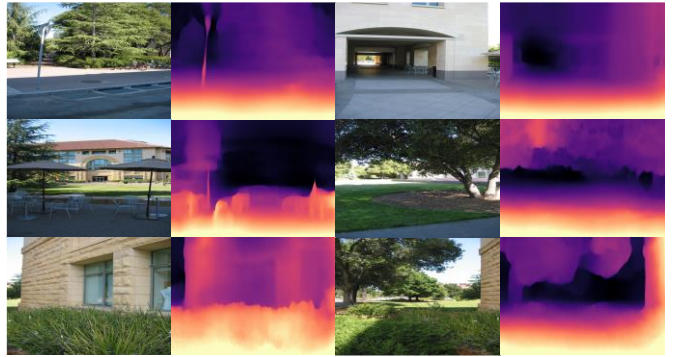

Fig. 3. An illustration of examples of depth predictions on the Make3D dataset. Note that our model is only trained on the KITTI dataset and directly tested on Make3D.

Xue et al. [37]) could be seen in Fig.2. It was shown that our method could reduce artifacts in low-texture regions of the image and improve the accuracy of close-range objects. The performance improvements mainly owe to the 3D geometric consistency is further verified by the 2D geometric constraints on the image planes, by alternative aligning depth with pose and aligning pose with depth can correct each other and the minimum loss instead of average loss is also useful for geometric constraints.

**Generalization ability on the Make3D dataset.** To illustrate the generalization ability of the trained model on other unseen datasets, we evaluated the network on the Make3D dataset, which was trained only by the KITTI dataset. Qualitative results were shown in Fig. 3. Despite the dissimilarities of the datasets, both in contents and camera parameters, we still achieved reasonable results.

**Pose Evaluation.** Although our method mainly concentrated on better depth estimation, we also evaluated the performance of relative pose estimation with competing methods on the official KITTI odometry benchmark using the Absolute Trajectory Error (ATE) metric over N-frame snippets (N=3 or 5), as in [11]. The pose estimation results in Table 2 showed the improvement over existing methods. We had observed that with the epipolar geometric constraints, the result of pose estimation would be notably improved, which is consistent with the report in [16].

### D. KITTI Ablation Study

**Performance of different losses.** To analyze the individual impact of each loss, we provided an ablation study over different combinations of losses. Models for depth and pose evaluation were trained only on KITTI raw dataset or odometry dataset respectively. We chose an incremental order for the proposed techniques to avoid too many loss combinations. As shown in Table 3, we had the following observations.

1) The 3D structure consistency was essential for improving the performance of depth estimation, while the log-scale 3D structural consistency loss could further improve the depth estimation. Our method was more stable overall all metrics, especially noticeable on metrics that were especially noticeable on metrics that sensitive to large depth errors, e.g. the Square Relative Error and RMSE.

2) Although the properties of epipolar geometry, low-rankness constraints and union-of-subspaces constraints, is

TABLE III
ABLATION STUDIES. EVALUATION OF DIFFERENT TRAINING LOSS CONFIGURATIONS AND TRAINING POLICIES. ALL MODELS ARE EITHER SOLELY TRAINED ON KITTI RAW DATASET WITHOUT PRE-TRAINING ON CITYSCAPES [5] AND POST-PROCESSING [11]. THE DEPTH ESTIMATION PERFORMANCE IS EVALUATED WITH MAXIMUM DEPTH CAPPED AT 80M. WHERE THE ERROR METRIC IS LOWER THE BETTER, AND THE ACCURACY METRIC IS HIGHER THE BETTER. THE BEST RESULTS OF EACH METRIC ARE BOLD.

| Loss Configuration and Alternative Training Manner | | | | | | Error metric | | | | Accuracy metric | | |
|---|---|---|---|---|---|---|---|---|---|---|---|---|
| Baseline | $\mathcal{L}_{3D}$ | $\mathcal{L}_{ep}$ | $\mathcal{L}_{lr}$ | $\mathcal{L}_{sub}$ | Alternate | Abs Rel | Sq Rel | RMSE | RMSE log | $\delta < 1.25$ | $\delta < 1.25^2$ | $\delta < 1.25^3$ |
| ✓ | - | - | - | - | - | 0.115 | 0.903 | 4.863 | 0.193 | 0.877 | 0.959 | 0.981 |
| ✓ | - | - | - | - | ✓ | 0.114 | 0.885 | 4.845 | 0.192 | 0.877 | 0.960 | 0.980 |
| ✓ | ✓ | - | - | - | - | 0.115 | 0.896 | 4.844 | 0.193 | 0.876 | 0.959 | 0.981 |
| ✓ | ✓(log) | - | - | - | - | 0.114 | 0.884 | 4.834 | 0.192 | 0.877 | 0.959 | 0.982 |
| ✓ | - | ✓ | - | - | - | 0.113 | 0.879 | 4.828 | 0.191 | 0.876 | 0.960 | 0.981 |
| ✓ | ✓(log) | ✓ | - | - | - | 0.113 | 0.872 | 4.825 | 0.189 | 0.882 | 0.960 | 0.981 |
| ✓ | ✓(log) | ✓ | ✓ | - | - | 0.114 | 0.870 | 4.812 | 0.192 | 0.876 | 0.960 | 0.981 |
| ✓ | ✓(log) | ✓ | ✓ | ✓ | - | 0.113 | 0.868 | 4.809 | 0.192 | 0.877 | 0.959 | 0.982 |
| ✓ | ✓(log) | ✓ | ✓ | ✓ | ✓ | **0.112** | **0.835** | **4.748** | **0.189** | **0.878** | **0.960** | **0.982** |

TABLE IV
ABLATIVE ANALYSIS OF THE GENERALIZATION OF OUR PROPOSED NETWORK ON VARIANT NETWORK STRUCTURES. ALL MODELS ARE SOLELY TRAINED ON THE MONOCULAR IMAGES OF KITTI RAW DATASET WITHOUT PRE-TRAINING ON CITYSCAPES [5] AND POST-PROCESSING [11].

| Networks | Error metric | | | | Accuracy metric | | |
|---|---|---|---|---|---|---|---|
| | Abs Rel | Sq Rel | RMSE | RMSE log | $\delta < 1.25$ | $\delta < 1.25^2$ | $\delta < 1.25^3$ |
| ResNet-18 | 0.112 | 0.835 | 4.748 | 0.189 | 0.878 | 0.960 | 0.981 |
| ResNet-50 | 0.107 | 0.792 | 4.661 | 0.182 | 0.887 | 0.962 | 0.983 |
| PackNet | 0.103 | 0.698 | 4.274 | 0.172 | 0.894 | 0.964 | 0.985 |

suitable for self-supervised optical flow estimation, the improvements for self-supervised depth estimation are limited.

In summary, the epipolar geometry helped both the pose and depth estimation, while the log-scale 3D geometric consistency terms also could improve the performance of depth estimation.

**Alternative Training Policy.** We also conducted an ablation study over the proposed alternative training policy compared with the jointly training policy. It showed that the alternative training policy was as effective as the jointly training policy and more effective by training with different geometric constraints.

**Different Depth Network Structures.** For the sake of completeness, as similar to [13], we also provided an ablative analysis of its generalization ability to different depth networks. To this end, we considered two variations on the well-performed structures, the Resnet50 [14] as the encoder and the Packnet [12]. The estimation results of this consideration were shown in Table 4, where we could see that the proposed method could consistently improve the performance with different depth networks, for all considered metrics.

## V. CONCLUSIONS

In this paper, we put forward a self-supervised depth and pose estimation architecture that incorporates both the geometric principles and the photometric-based learning metrics. Our main contribution is to better utilize the mutual dependency of the depth and pose learning by their alternative training with different geometries and simplify the ICP registration based optimization by incorporating epipolar geometry. Also the log-scale 3D structural consistency loss and the epipolar geometry embedded ICP registration are adopted in the respective tasks. To make the result more robust and reliable, we incorporate novel ingredients by the properties of epipolar geometry, namely the low-rankness and self-expression in union-of-subspaces constraints, for depth and pose networks learning respectively. Our method tries to eliminate negative effects of possible objects movement by self-expression in the union of epipolar subspaces. The experimental results also demonstrate that our method can obtain depth maps with better contour for the foreground target and can generalize well to the unseen datasets. Further explorations include enforcing consistency across the whole dataset and applying to the uncalibrated dataset. The weights of variety of loss functions are set by the experience values and trial and error, various adaptive weighting could be considered in the future.